\documentclass[a4paper,twoside]{article}

\usepackage[letterpaper]{geometry}

\usepackage{amsfonts,amsmath,graphicx,booktabs}
\usepackage{algorithm2e}
\usepackage{balance}
\usepackage{epsfig}
\usepackage{subcaption}
\usepackage{calc}
\usepackage{amssymb}
\usepackage{amstext}
\usepackage{amsmath}
\usepackage{amsthm}
\usepackage{multicol}
\usepackage{pslatex}
\usepackage{apalike}
\usepackage{SCITEPRESS}     
\SetKwRepeat{Do}{do}{while}%

\begin{document}

\title{Med2Meta: Learning Representations of Medical Concepts with Meta-Embeddings}

\author{\authorname{Shaika Chowdhury\sup{1}, Chenwei Zhang \sup{2}, Philip S. Yu\sup{1} and Yuan Luo\sup{3}}
\affiliation{\sup{1}Department of Computer Science, University of Illinois at Chicago, Chicago, Illinois}
\affiliation{\sup{2} Amazon, Seattle, Washington}
\affiliation{\sup{3} Department of Preventive Medicine, Northwestern University, Chicago, Illinois}
\email{\{schowd21, psyu\}@uic.edu, cwzhang@amazon.com, yuan.luo@northwestern.edu}
}



\abstract{Distributed representations of medical concepts have been used to support downstream clinical tasks recently. Electronic Health Records (EHR) capture different aspects of patients' hospital encounters and serve as a rich source for augmenting clinical decision making by learning robust medical concept embeddings. However, the same medical concept can be recorded in different modalities (e.g., clinical notes, lab results) --- with each capturing salient information unique to that modality --- and a holistic representation calls for relevant feature ensemble from all information sources. We hypothesize that representations learned from heterogeneous data types would lead to performance enhancement on various clinical informatics and predictive modeling tasks. To this end, our proposed approach makes use of \textit{meta-embeddings}, embeddings aggregated from learned embeddings. Firstly, modality-specific embeddings for each medical concept is learned with graph auto-encoders. The ensemble of all the embeddings is then modeled as a meta-embedding learning problem to incorporate their correlating and complementary information through a joint reconstruction. Empirical results of our model on both quantitative and qualitative clinical evaluations have shown improvements over state-of-the-art embedding models, thus validating our hypothesis. }

\onecolumn \maketitle \normalsize \setcounter{footnote}{0} \vfill

\section{Introduction}

With the increase in healthcare efficiency and improvement in patient care that comes with archiving medical information digitally, more healthcare facilities have started adopting Electronic Health Records (EHRs) \cite{shickel2018deep,knake2016quality,charles2013adoption}. EHRs store a wide range of heterogeneous patient data (e.g., demographic information, unstructured clinical notes, numeric laboratory results, structured codes for diagnosis, medications and procedures) as a summary of the patient's entire hospital stay. As a result of the richness of medical content that can be mined from EHR, its use in predictive modeling tasks in the medical domain has become ubiquitous. 

How good a machine learning algorithm performs is dependant on the representations of the data that it is able to learn \cite{bengio2013representation}. However, the high dimensionality, diversity of data and complex associations between clinical variables in EHR means the intermediate data representation learning task is not trivial. Feature engineering with the help of a domain expert was prevalent in earlier works to specify which clinical variables from the EHR to consider as the input features \cite{jensen2012mining}. Although the features could be more precise as hand-picked with domain knowledge, the manual effort, scalability issues and inability to generalize render such approach undesirable \cite{miotto2016deep}. 

Distributed representations of words, known as word embeddings, have brought immense success in numerous Natural Language Processing (NLP) tasks \cite{goldberg2016primer}. Recent works in deep learning applications for EHR \cite{choi2016learning,choi2016medical,tran2015learning,choi2016multi,choi2017gram} emulated the concept of word embedding as learning vector representations of medical concepts. They focused on learning the embeddings from predominantly one modality (e.g., unstructured notes or structured clinical events) that excludes relevant information about medical concepts found in other modalities. For example, a patient's symptoms for a disease can be mentioned in physician notes but could be missing from their structured clinical event data of clinical codes.  However, trying to fuse information from different modalities in EHR presents the following obstacles for representation learning,

1.	\textbf{Inconsistency in medical concept terminology}. In structured clinical events, the medical concept is represented with ICD-9/ICD-10 clinical code. While in unstructured clinical notes, it is either mentioned with a formal medical term for the concept or an informal analogous term/phrase implicitly. It is difficult to consistently detect the presence of a medical concept across different modalities.

2. \textbf{Varying contexts}. Prediction based embedding approaches \cite{choi2016medical,choi2016multi} work well on structured modality where each patient can be represented as a sequence of visits of codes, and can consider context in terms of the other neighboring codes within the same visit. However, for unstructured clinical notes, the context can be noisy due to the presence of text describing all aspects of a patient's admission (e.g., past medical history).

3. \textbf{Feature Associations Complexity}. Other types of patient information such as demographics and laboratory results are possibly important signals in prediction tasks. However, modeling their non-linear and complex relations with the medical concepts is not straightforward.

4. \textbf{Interpretability}. The resulting learned embedding should be understandable as clinicians want to know the underlying reasons for predictive results, that should also comply with medical knowledge.

To address these obstacles our proposed approach, Med2Meta, is formulated as a meta-embedding learning problem. \textit{Meta-embeddings} are embeddings obtained from the ensemble of existing embeddings. As embeddings vary in terms of the corpus they are created from and the approach used to learn them, \cite{chen2013expressive,yin2015learning} have found that they capture different semantic characteristics. Meta-embedding learning exploits this idea and combines the semantic strengths of different types of embeddings, which has been shown to outperform single embedding on tasks such as word similarity and analogy \cite{yin2015learning}. Med2Meta differs from traditional meta-embedding learning in that it does not combine existing pre-trained embeddings to obtain the new embedding for each medical concept, as most of these models are designed for one data modality. Rather, feature-specific embedding for each medical concept in the EHR is first learned with graph auto-encoder by considering each heterogeneous data type as a different view. Using graph auto-encoder for learning embeddings gives us the benefit of being able to model the relations between different types of medical concepts through the graph's structure and, at the same time, infuse relevant feature information for each medical concept collected from a particular modality. In particular, with each medical concept modeled as a node in the graph and edges between two nodes signifying corresponding relationship found in EHR data w.r.t a modality, embeddings are learned by considering features extracted from that modality as a separate view. Med2Meta considers three different heterogeneous data types in EHR as separate views --- \textit{demographic information} (\textit{dem}), \textit{laboratory results} (\textit{lab}) and \textit{clinical notes} (\textit{notes}). Each graph is constructed with a novel graph construction approach, where unique medical concepts from structured clinical events are extracted to serve as the nodes of the graph.

The embeddings learned from each view are then considered as input sources for the meta-embedding process, which are fused together using Dual Meta-Embedding Autoencoders (Dual-MEAE). The encoders in Dual-MEAE are inspired by the idea in  \cite{roweis2000nonlinear}, where each data point and its neighbors can be expected to lie on or close to a locally linear patch of the manifold, where similar instances should result in similar positions in the embedding space. Therefore in Dual-MEAE, a pair of encoders are allocated for each view --- one for the source embedding and the other for the average embedding of the most similar medical concepts based on that view ---- in order to project different modalities to a common meta-embedding space. That is, the embedding of a medical concept can be expected to be semantically similar to that of its most similar medical concepts and, hence, should also lie nearby in the meta-embedding space. To enable this, a single decoder jointly tries to reconstruct the original input source and the average embedding from the latent meta-embedding representation. By minimizing the reconstruction errors across all the views in an unified manner, the intermediate meta-embedding representation is able to retain correlating information among medical concepts within the same view, as well as capture complementary information across different views, hence learning a holistic vector representation for each medical concept. In general, Med2Meta's contributions are threefold,

$\bullet$ Learns holistic embeddings of medical concepts that fuses information from heterogeneous data types in EHR through a meta-embedding process. 

$\bullet$ Modality-specific embeddings capture the semantic features of each EHR data type and are generated using Graph Auto-Encoders. 

$\bullet$ Outperforms current state-of-the-art models in both qualitative and quantitative experiments for a publicly available EHR dataset.


\begin{figure}[ht!]
    \centering
    \includegraphics[width=0.9\linewidth]{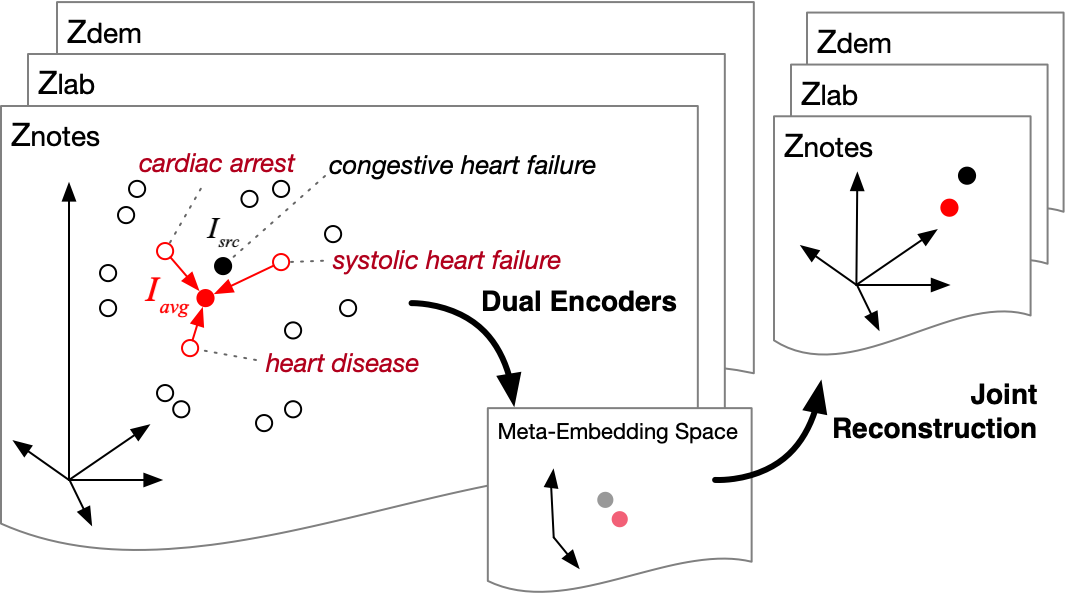}
    \caption{Proposed Dual-MEAE Model: Fuses embeddings from different views with dual encoders for each view and a single decoder to reconstruct jointly across all views.}
    \label{fig::Figure_2}
\end{figure}

\begin{algorithm}[ht!]
  \SetAlgoLined
  \KwData{$I^{i}_{src}$, $I^{i}_{avg} \in \mathbb{R}^{V \times d'}$ for $i \in \{dem, lab, notes\}$}
  
  Load all $I^i_{src}$ and $I^i_{avg}$ \;
  \Repeat{convergence}{
  \ForEach{medical concept $c \in V$}{
  \ForEach{view $i \in k$}{
  Feed $I^i_{src}(c)$ into ${Enc}^i_{src}(c)$ to get output $T^i_{src}(c)$ \;
  Feed $I^i_{avg}(c)$ into ${Enc}^i_{avg}(c)$ to get output $T^i_{avg}(c)$ \;
  Concatenate $T^i_{src}(c)$ and $T^i_{avg}(c)$ to get $T^i(c)$ \;
  Pass $T^i(c)$ through a dense layer to get $m^i(c)$ \;
  }
  Concatenate all $m^i(c)$ to get meta-embedding, $m(c)$ \;
  Feed $m(c)$ into the decoder, ${Dec}$, to get $\Dot {Dec}(c)$ \;}
  Calculate total loss $\mathcal{L}$ in eq. 6 \;
  Update parameters according to the gradient of $\mathcal{L}$ \;
  }
  \caption{Dual-MEAE Optimization Algorithm}\label{alg:the_alg}
\end{algorithm}

\section{Meta-Embedding Problem Definition}
Let $S^i \in \mathbb{R}^{{V^i} \times {d^i}}$ represent the $i$-th source medical concept embedding, where $i\in \{1,\dots k\}$, for total $k=3$ source embeddings representing the three different views in our case. Here $V^i$ is the medical concept vocabulary covered by the $i$-th source embedding. In our case, $V^i$ for all three sources $i\in \{1,\dots k\}$ are the same as the total number of unique medical concepts, \textit{V}, used to create each modality-specific graph is the same, and hence we consider $V^i$ = \textit{V}. The dimension of medical concept embedding for the respective source is denoted by $d^i$,  where we set all $d^i$ to be of the same dimension, $d$. The meta-embedding learning problem is then defined as learning an embedding $m(c)\in \mathbb{R}^b$ of dimension $b$ in the meta-embedding space for each medical concept $c \in V$.

\section{Med2Meta Architecture}

Our proposed approach comprises two steps: \textit{1) obtaining  modality-specific embeddings} and \textit{2) meta-embedding learning}. In step 1, Graph Auto-Encoder is trained on each modality features in turn to generate feature-specific embeddings to be considered as input sources into step 2. While in step 2, all the input source embeddings are ensembled through meta-embedding learning to obtain the new embedding for each medical concept. 

\subsection{Generating Modality-Specific Embeddings}
The embedding learned for each medical concept should holistically capture relevant features from each data modality in EHR, as patient information coverage is not the same across all the modalities. For instance, information regarding symptoms experienced during a disease, its severity and other important observations made by physicians/ nurses are probably only found in clinical notes. Hence, learning the embeddings only from structured clinical records is a sub-optimal approach. We use the same Graph Auto-Encoder technique as used in \cite{chowdhury2019mixed}, however, in this work graph for each view is constructed using EHR information collected from all the patients.

\subsection{Fusion through Meta-Embedding Learning}
The three types of modality-specific embeddings obtained through graph auto-encoder exist in different vector spaces and capture semantic relations of the respective view. In order to learn an embedding that collectively reflects semantic associations across different modalities in EHR for each medical concept, semantic knowledge from all the feature-specific embeddings need to be integrated. Performing the fusion by casting it as a meta-embedding problem through autoencoding is meaningful as we intend to learn the ensembled embedding as a meta-embedding by reconstructing the modality-specific embeddings jointly. The projection to a common meta-embedding space would therefore lead to coherent enforcement of mutual and correlative information present in the embeddings. The proposed Dual-MEAE model is depicted in Figure \ref{fig::Figure_2}.

We consider each modality specific embedding, extracted from a view in step 1, as an input source embedding, $S^i$, into Dual-MEAE. In each source embedding space, medical concepts that are semantically similar in that view would exist geometrically closer and thus can be expected to have similar distributed vectors. So, our proposed approach reconstructs each medical concept jointly from both the source embedding and average embedding of its most similar medical concepts.  

Each component of our meta-embedding model, Dual-MEAE, is discussed in detail in the following sub-sections.

\subsubsection{Encoder}
On the encoder side, a set of dual encoders, ${Enc}^i_{src}$ and ${Enc}^i_{avg}$, is set aside for each view \textit{i} from $i\in \{dem, lab, notes\}$. For each medical concept \textit{c}, its learned source embedding with respect to view \textit{i} and the average of the source embeddings of most similar medical concepts to $c$ in that view are fed as inputs, $I^i_{src}(c)\in \mathbb{R}^d$ and $I^i_{avg}(c)\in \mathbb{R}^d$, into ${Enc}^i_{src}$ and ${Enc}^i_{avg}$ respectively. That is, the output feature matrix $Z^{i}\in \mathbb{R}^{V \times d}$ obtained for each view from step 1 is considered as the respective source embedding, $S^i$, for that view. 
Henceforth, $Z^{i}[c, :]$, which is the $c$-th row of $Z^{i}$ for $c\in V$ and represents the GAE-generated distributed embedding vector of $c$-th medical concept, is fed as input $I^i_{src}(c)$ into ${Enc}^i_{src}$ for the respective view. 
From our preliminary results, we found that projecting the output feature matrix from dimension ${N \times d} \rightarrow V \times d'$ where $d' < d$ through principal component analysis (PCA) \cite{wold1987principal} improves performance; so we feed embedding vector extracted from the PCA output matrix $Z^{i'}$ instead. 

On the other hand, the accumulation of the source embeddings of the most similar medical concepts to \textit{c} with respect to view \textit{i} is computed as their average, $avg^i$(c), and is considered as the input, $I^i_{avg}(c)$, to the second encoder, ${Enc}^i_{avg}$. We specifically chose to compute the average of the neighborhood embeddings as it has shown to perform well in recent studies \cite{coates2018frustratingly}. To find the most similar medical concepts of concept \textit{c}, the top three medical concepts sorted by highest values of dice coefficient for that view computed in step 1 are selected. Top three gave us best results based on hyperparameter analysis on validation set.

The dual encoders encode their inputs $I^i_{src}$ and $I^i_{avg}$, so that latent attributes and their non-linear relationships could be effectively learned. Let us call the encoded representations from the dual encoders for each medical concept as their target embeddings, $T^i_{src} \in \mathbb{R}^{d'}$ and $T^i_{avg} \in \mathbb{R}^{d'}$, defined as

\begin{equation}
\resizebox{.48\hsize}{!}{$T^i_{src}(c) = {Enc}^i_{src}(I^i_{src}(c))$}
\end{equation}

\begin{equation}
\resizebox{.48\hsize}{!}{$T^i_{avg}(c) = {Enc}^i_{avg}(I^i_{avg}(c))$}
\end{equation}

Each encoder is implemented as a fully connected neural network layer, with $d'$ hidden units and ReLU \cite{nair2010rectified} activation.


\subsubsection{Meta-Embedding Space} 
We first concatenate the respective encoded input representations as $T^i(c)\in \mathbb{R}^{2d'}$,

\begin{equation}
\resizebox{0.5\hsize}{!}{$T^i(c) = [T^i_{src}(c)$; $T^i_{avg}(c)]$}
\end{equation}

here [] denotes concatenation.

To get meta-embedding ${m(c)}\in \mathbb{R}^b$, each $T^i(c)$ is first passed through one dense layer to model it to respective intermediate representation, $m^i(c)\in \mathbb{R}^{2d'}$. The concatenation of all $m^i(c)$ is considered as the meta-embedding $m(c)$, defined as, 

\begin{equation}
\resizebox{0.33\hsize}{!}{$m(c) = [m^i(c)]$}
\end{equation}

for $i \in \{dem, lab, notes\}$ and dimension $b = 6d'$.

As Dual-MEAE jointly encodes and decodes across all the views, the meta-embedding is updated with view-specific feature information in a meta-embedding space. Thus, the meta-embedding space will eventually be able to capture a holistic semantics comprising of associations between medical concepts within same as well as different views.

\subsubsection{Decoder}
In order to decode the semantically composite meta-embedding representation, a single decoder \textit{Dec} maps the latent representation, $m(c)$, from the meta-embedding space back to corresponding source and average embeddings across all views simultaneously. The decoder is implemented as a neural network of dense layers, each with $d'$ hidden units and ReLu activations. The output of the decoder is defined as, 

\begin{equation}
{\dot {Dec}(c)  = \textit{Dec}(\textit{m(c)})}.
\end{equation}

$\dot {Dec}(c) \in \mathbb{R}^{6 \times d'}$ contains the reconstructed source and average embeddings for the three views.

This joint decoding from the meta-representation simultaneously across all the views drives Med2Meta in embedding both mutually related and complementary information of heterogeneous data types into the meta-embedding space. 

\subsubsection{Objective Function}
The meta-representation, $m(c)$, optimized through the minimization of the reconstruction losses across all the views is taken as the final embedding for the corresponding medical concept. The overall objective function for total $k$ views and $V$ medical concepts in the training set, outlined in Algorithm \ref{alg:the_alg}, is defined as,


\begin{equation}
\begin{aligned}
\mathcal{L} = &\sum\limits_{c \in V}{}\sum\limits_{i \in k}(  
{\omega _1}{{\left \| {T_{src}^i(c)  - T_{avg}^i(c)} \right\|}^2} +\\
& {\omega _2}{{\left\| {\dot {Dec}(c)[i_{src},:] -  I_{src}^i(c)} \right\|}^2} +  \\ 
& {\omega _3}{{\left\| {\dot {Dec}(c)[i_{avg},:] - I_{avg}^i(c)} \right\|}^2})
\end{aligned}
\end{equation}

$\dot {Dec}(c)[i_{avg},:]$ and $\dot {Dec}(c)[i_{src},:]$ refer to the decoded embeddings for the average and source embeddings respectively of view $i$ for medical concept $c$. The first term in the loss is responsible for infusing common associations between a medical concept and other similar medical concepts, whereas, the second and third try to preserve information essential locally with respect to each view for each medical concept during the respective reconstructions. Moreover, the degree of these properties in the loss can be controlled through the values of the coefficients $\omega$. 

\section{Experimental Setup}
\subsection{Source of Data}
We evaluate our model on the publicly available MIMIC-III dataset \cite{johnson2016mimic}. This database contains de-identified clinical records for $>$ 40K patients admitted to critical care units over 11 years.  ICD-9 codes for diagnosis/procedures and NDC codes for medications were extracted from patients with at least two visits to construct each graph. Table \ref{tab::Table1}  outlines other statistics about the data.

\begin{table}[ht!]
\centering
\caption{Data Statistics Summary of MIMIC-III
}
\label{tab::Table1}
\begin{tabular}{llll}
\toprule
                     & 	MIMIC-III\\ \midrule
\# of patients                 & 7,499  \\
\# of visits                & 19,911  \\ 
avg. \# of visits per patient     & 2.66\\ 
\# of unique ICD9 codes  & 4,893 \\
avg. \# of codes per visit  & 13.1 \\
max \# of codes per visit  & 39 \\
\bottomrule
\end{tabular}
\end{table}

\begin{table*}[]
\centering
\caption{Performance of different embeddings on three different tasks.}
\label{tab::Table2}
\begin{tabular}{l|lll|lll|lll}
\toprule
                    & \multicolumn{3}{l}{Heart Failure \textbf{(HF)}
Prediction

} & \multicolumn{3}{l}{Relation \textbf{(Rel)} Classification
} & \multicolumn{3}{l}{Semantic \textbf{(Sem)}} \\
Metrics             & AUC-ROC       & Accuracy         & AUC-PR         & AUC-ROC     & Accuracy           & AUC-PR          & Similarity $\rho$         \\ \midrule
M2M  & $\textbf{0.685}$       & $\textbf{0.638 }$    & $\textbf{0.674}$       & $\textbf{0.833}$       & $\textbf{0.950}$      & $\textbf{0.967}$     & $\textbf{0.650}$   \\
$M2M\_d$       & 0.640     & 0.619     & 0.635     & 0.346       & 0.500      & 0.692      & 0.154      \\
$M2M\_l$  & 0.627     & 0.618     & 0.629     & 0.354       & 0.880      & 0.923      & 0.577      \\
$M2M\_n$  & 0.620     & 0.616     & 0.630     & 0.352       & 0.880      & 0.750      & 0.154      \\
$M2M\_{s}$  & 0.657     & 0.628     & 0.632     & 0.481       & 0.227      & 0.385      & 0.154      \\
\hline
CONC  & 0.666     & 0.626     & 0.673     & 0.481       & 0.500      & 0.692      & 0.154      \\
AVG  & 0.639     & 0.627     & 0.634     & 0.370       & 0.700      & 0.846      & 0.327      \\
SVD  & 0.679     & 0.605     & 0.673     & 0.574       & 0.550      & 0.615      & 0.154      \\
Hot  & 0.520     & 0.587     & 0.587     & 0.426       & 0.570      & 0.077      & 0.119      \\
GV  & 0.592     & 0.589     & 0.575     & 0.500       & 0.500      & 0.692      & 0.153  \\   
M2V  & 0.678     & 0.577     & 0.670     & 0.648      & 0.750      & 0.850      & 0.576      
\\\bottomrule
\end{tabular}
\end{table*}

\subsection{Evaluation Tasks and Metrics}
The performance of Med2Meta embeddings is demonstrated using both quantitative and qualitative evaluations. 
\subsubsection{Quantitative}
The embeddings are evaluated on the following three medical tasks,

\textbf{Semantic Similarity Measurement}: The semantic similarity between two medical concepts is calculated as the cosine similarity between their learned embeddings and Spearman Correlation Coefficient, $\rho$, measured against manual rating based on whether two concepts fall under same hierarchical grouping of ICD-9 codes collected from clinical classification software (CCS) \cite{elixhauserclinical}.

\textbf{Relation Classification}: We consider two types of relation, \textit{MAY-TREAT} and \textit{MAY-PREVENT}, from the National Drug File Reference Terminology (NDF-RT), which is an ontology containing sets of relations occurring between drugs and diseases. \textit{MAY-TREAT} relation holds for drug-disease pairs where a drug may be used to treat a disease. While \textit{MAY-PREVENT} forms a relation between a drug and a disease if the drug may be used to prevent the disease. A 2-nearest neighbor classifier is trained on relation tuples from triples $($\textit{rel}, \textit{c1}, \textit{c2}$)$ where $rel\in \{MAY-TREAT, MAY-PREVENT\}$. 
That is, for each test tuple, the cosine similarity between the vector offset between the drug-disease embeddings involved (i.e., $m(c1)$ - $m(c2)$) and that of all the other tuples in the dataset are first computed. Then the cosine similarities are ranked in descending order, and the evaluation metrics (Accuracy, AUC-ROC and AUC-PR) are measured. If any of the two top-ranked tuples holds the same relation as the test tuple, it is considered as a correct match.

\textbf{Outcome Prediction}: This is a binary prediction task that tries to predict whether the patient is at risk of developing a disease in the future visit, $v_{t+1}$, trained on visit embedding sequence up to $v_t$. We focus on patients with heart failure (HF) disease. Thereby, we examine only patients with at least two visits and check if they contain an occurrence of heart failure in their $v_{t+1}$ visit. These are considered as the instances belonging to the positive class (HF). As average number of visits per patient is 2 in MIMIC-III, $v_{t+1}$ in our case is 1. A binary logistic regression classifier is trained/tested to perform this prediction task. 
The train/test/validation split for positive instances 
is 75\%/12.5\%/12.5\%
and the same split is applied to equal number of total negative instances, where negative instances are formed with patients who do not have HF code in their record up to the $v_{t+1}$ visit.

\subsubsection{Qualitative}
We use t-Distributed Stochastic Neighbor Embedding (t-SNE) \cite{maaten2008visualizing}, which is a visualization technique that maps data in a high-dimensional space to two or three dimensions, to examine if natural clusters of medical codes contain similar diagnosis, medication and procedure concepts.

\subsection{Baseline Models}
To show that the contribution of inclusion of heterogeneous data and meta-embedding learning approach in Med2Meta leads to superior empirical results, we compare against Med2Meta (\textbf{M2M}) trained on single view as well as other vector ensemble methods which include,

\textbf{M2M\_d}: Proposed approach with embeddings learned from graph auto-encoder with only demographics features.

\textbf{M2M\_l}: Proposed approach with embeddings learned from graph auto-encoder with only laboratory test results features.

\textbf{M2M\_n}: Proposed approach with embeddings learned from graph auto-encoder with only clinical notes features.

\textbf{M2M\_s}: Proposed approach where on the encoder side instead of dual encoders, only a single encoder is considered for each view that takes as input the source embedding, $I^i_{src}$.

\textbf{CONC}: The source feature-specific embeddings are simply concatenated to represent the final embedding of each medical concept. We $\ell{2}$ normalize each source feature-specific embedding before concatenation to ensure that each embedding contributes equally during the similarity computation.

\textbf{AVG}: The source feature-specific embeddings are averaged to represent the final embedding of each medical concept. $\ell{2}$ normalization is performed on each source feature-specific embedding before averaging.

\textbf{SVD}: Consider a matrix $C$ of dimension $N \times {d_{svd}}$, where $d_{svd}$ is dimension of the resulting embedding from the concatenation of the $\ell_{2}$ normalized embeddings for each medical concept. Singular Value Decomposition (SVD) \cite{golub1970singular} is applied on $C$ to get the decomposition \textit{C} = $USV^T$. For each concept, the corresponding vector in \textit{U} is considered as the SVD embedding.

 We also compare against hot vector representation and two state-of-the-art embedding models,

\textbf{Hot}: This is the one-hot vector representation of a concept $c$, $v_c \in \{0,1\}^N$. Only the dimension corresponding to the concept is set to 1. 

\textbf{GV}: GloVe \cite{pennington2014glove} is an unsupervised learning approach of word embeddings based on word co-occurrence matrix.

\textbf{M2V}: Med2Vec \cite{choi2016multi} is a two-layer neural network for learning lower dimensional representations for medical concepts.

\begin{figure*}[ht!]
    \begin{subfigure}[t]{0.25\linewidth}
        \centering
        \includegraphics[width=\linewidth]{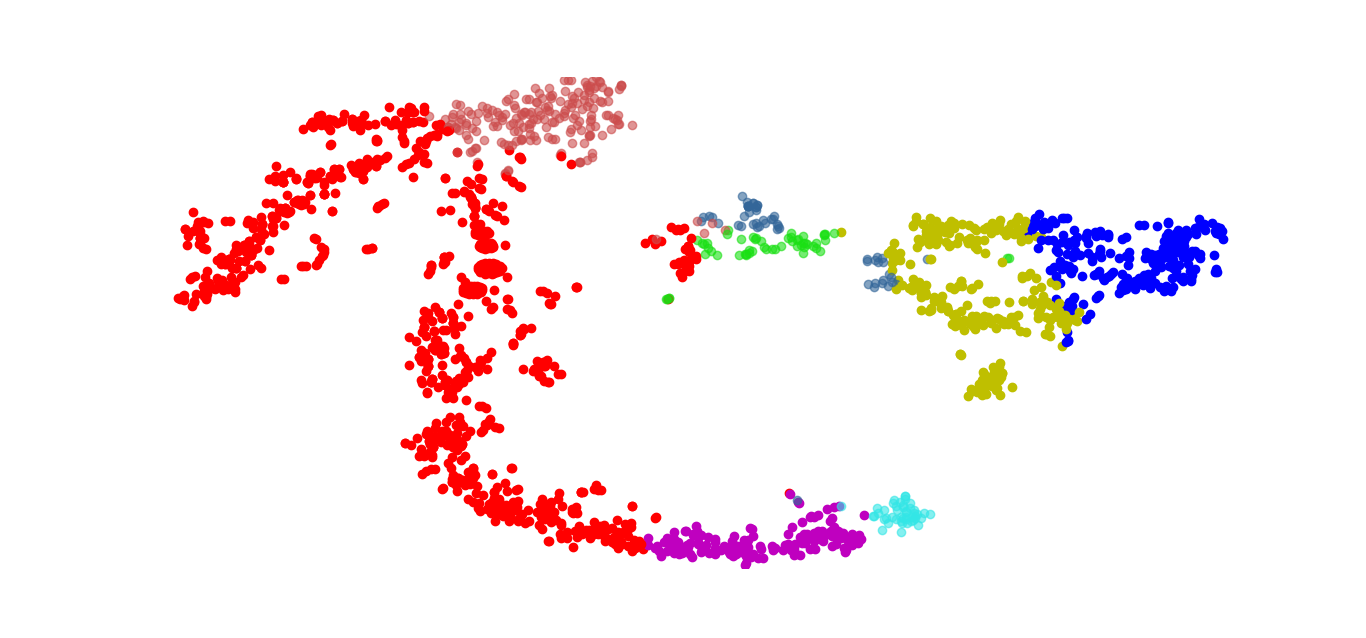}
        \caption{Med2Meta}\label{fig::Figure_new_my}
    \end{subfigure}%
    \begin{subfigure}[t]{0.25\linewidth}
        \centering
        \includegraphics[width=\linewidth]{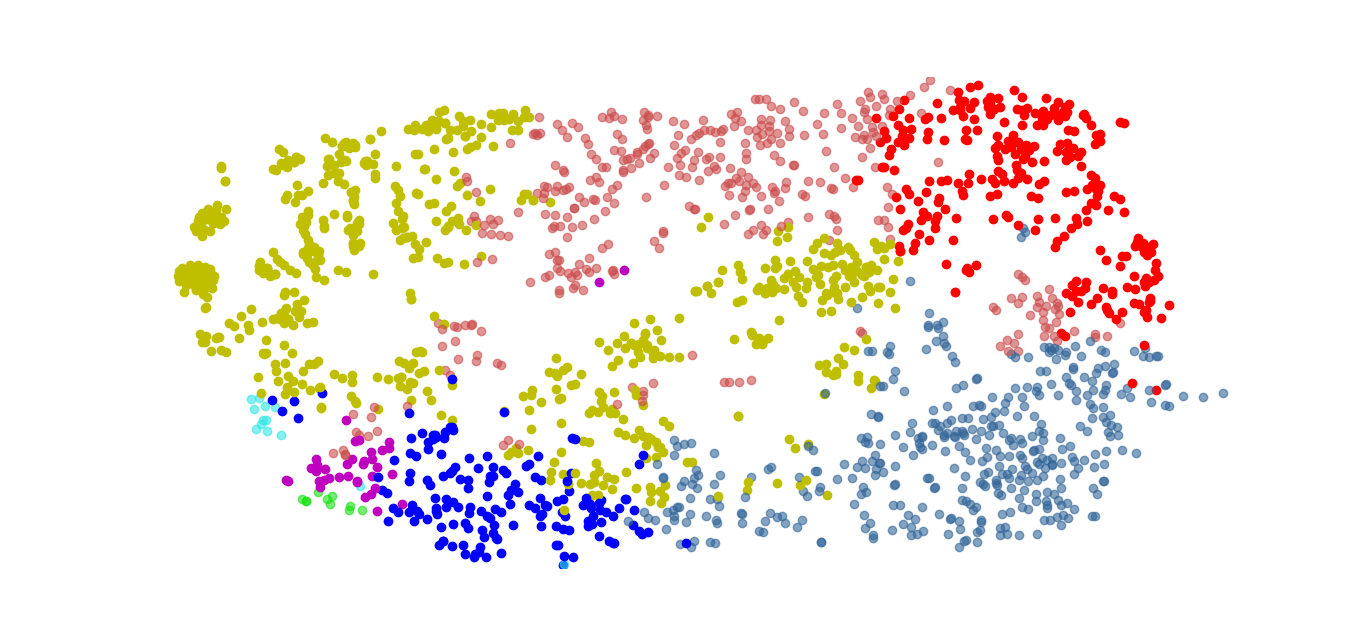}
        \caption{CONC}\label{fig::Figure_new_conc}
    \end{subfigure}%
    \begin{subfigure}[t]{0.25\linewidth}
        \centering
        \includegraphics[width=\linewidth]{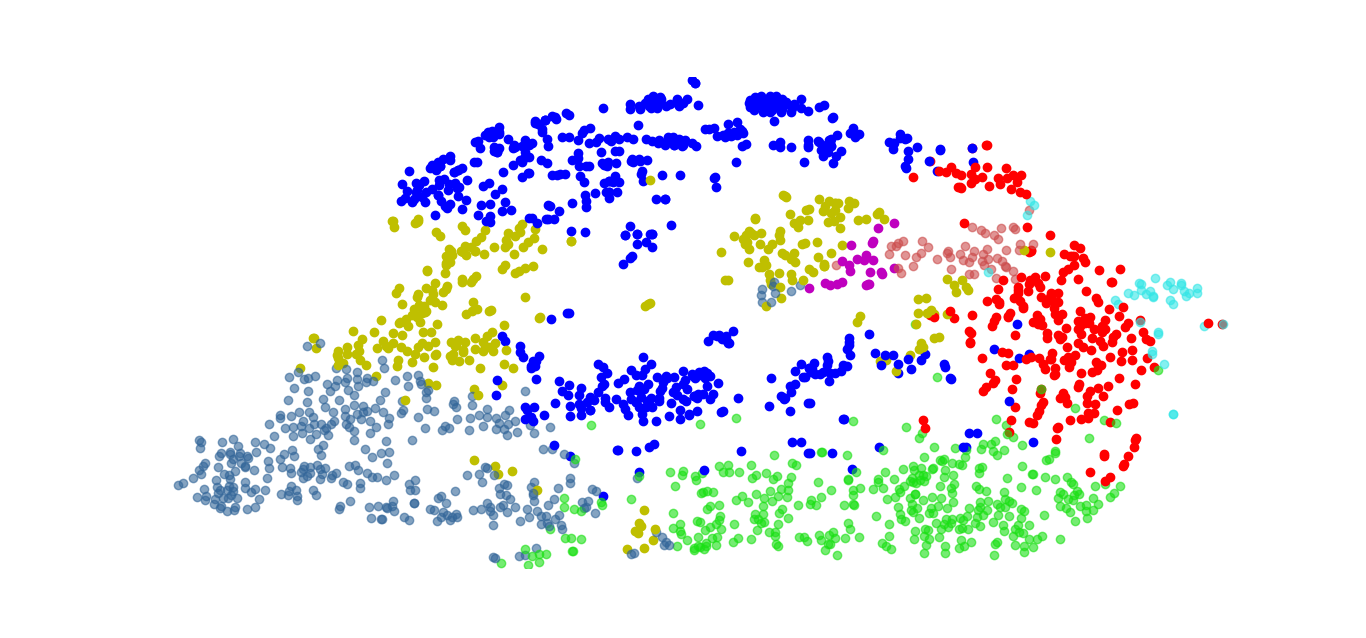}
        \caption{AVG}\label{fig::Figure_new_avg}
    \end{subfigure}%
    \begin{subfigure}[t]{0.25\linewidth}
        \centering
        \includegraphics[width=\linewidth]{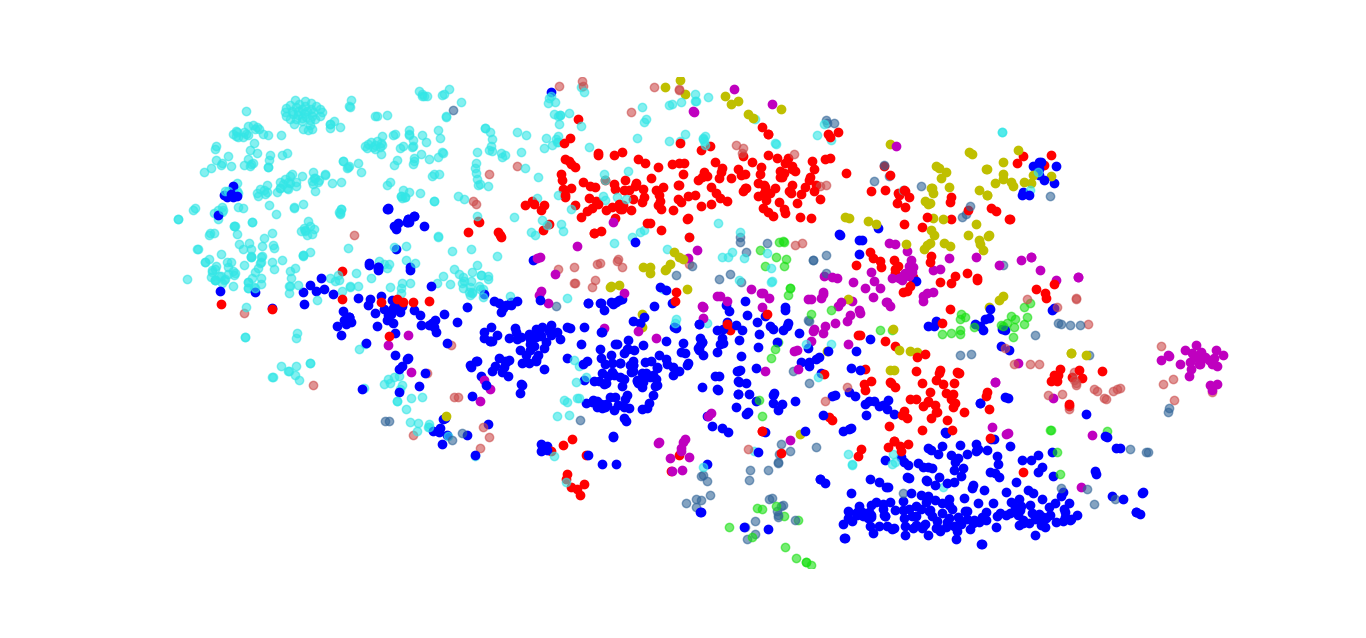}
        \caption{SVD}\label{fig::Figure_new_svd}
    \end{subfigure}%
    \caption{The t-SNE plots of learned embedding spaces. The color of the dot in the Figure indicates the cluster the medical concept has been assigned to by K-means Clustering.}
\end{figure*}

\section{Experimental Results}
\subsection{Quantitative Analysis}

Table  \ref{tab::Table2} reports results of performance of embeddings obtained with different models on heart failure (HF) prediction, semantic similarity (Sem) between medical concepts and relation classification (Rel) tasks. For heart failure prediction and relation classification tasks, the embeddings are evaluated in terms of AUC-ROC, AUC-PR and Accuracy, and for semantic similarity with Spearman Correlation Coefficient ($\rho$). We see that our multi-view, meta-learning approach Med2Meta (M2M) outperforms the single view models $M2M\_d$, $M2M\_l$ and $M2M\_n$ on all the tasks. This reinforces the contribution of learning embeddings from multi-modal data and means that different types of embeddings contribute significantly according to their semantic strengths. Among the single view models, surprisingly, $M2M\_d$ is seen to perform better than the other two in HF task. This could be attributed to most HF patients having distinctive demographics (e.g., older patients).    

The benefit of having dual encoders and reconstructing jointly from the source embedding and average of most similar medical concepts for each view can be seen when comparing $M2M$'s performance against $MSM\_s$, which is ablated version of $M2M$ with a single encoder for each view. This indicates that the dual encoder encourages the consolidated feature space to have locally linear patches for instances with similar semantics. 

Compared to the other types of fusion approaches, fusing the source embeddings through meta-embedding learning in M2M leads to considerable AUC-ROC gain on all tasks. AVG performing at par with SVD and CONC, aligns with findings in \cite{coates2018frustratingly}, which found AVG to outperform them in several benchmark NLP tasks. With regard to M2M's superior performance gains, we can conclude that the joint reconstruction of the source embedding and its most similar concept embedding aggregation across all the views is able to integrate both local semantic information in terms of the closest medical concepts and global semantic information in terms of multi-modality.

Rows 9-11 in Table  \ref{tab::Table2} show that M2M outperforms simple representation, one-hot vector, and state-of-the-art embedding models, GloVe and Med2Vec, on all the tasks. M2M performs comparably to Med2Vec in heart failure prediction task and exceptionally by 29\% and 13\% increases for relation classification and semantic similarity tasks respectively. Although Med2Vec also includes demographic information during embedding learning, it does not learn from different modalities (e.g., lab results, clinical notes) which can contain salient information for predictive tasks - as is confirmed by better performance of M2M across all the tasks.

\subsubsection{Varying Training Data Size}
\begin{figure}[ht!]
    \centering
    \includegraphics[width=\linewidth]{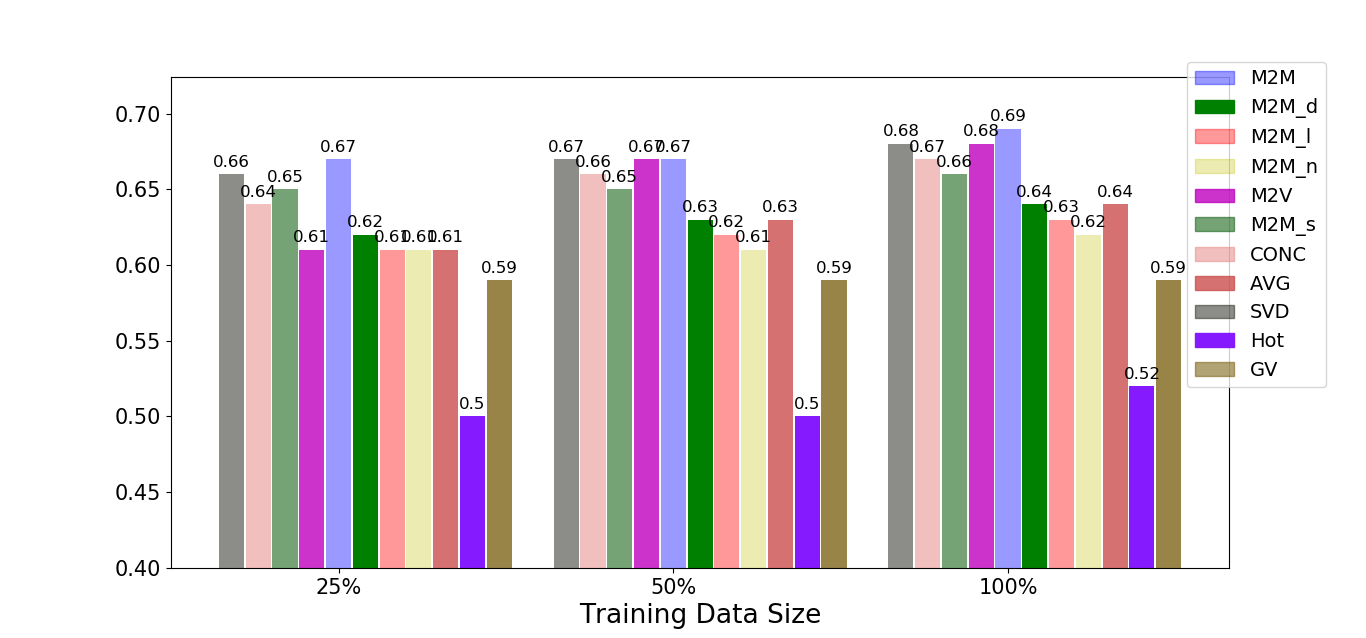}
    \caption{Performance (i.e., AUC-ROC) on HF Prediction task with varying data size.}
    \label{fig::Figure_3}
\end{figure}

To check how different types of embedding perform on smaller training data, we randomly sub-sample instances from the complete training dataset to create new datasets. Figure \ref{fig::Figure_3} depicts their heart failure prediction performance, in terms of AUC-ROC score, on being trained on 25\%, 50\% and 100\% sized training data. It can be seen that M2M's performance even under data insufficiency is above all the baselines, demonstrating its effectiveness for predictive modeling in the medical domain.

\subsection{Qualitative Analysis}
Our embeddings are qualitatively assessed by visualizing their t-SNE plots in 2-D space shown in Figure \ref{fig::Figure_new_my}. All the medical concepts are first grouped into clusters using K-means Clustering. The color of the dot in Figures \ref{fig::Figure_new_my}--\ref{fig::Figure_new_svd}  indicates the cluster the medical concept has been assigned to. For comparison, t-SNE plots of embeddings obtained by the baselines CONC, SVD and AVG are also shown in Figures \ref{fig::Figure_new_conc}, \ref{fig::Figure_new_avg}, \ref{fig::Figure_new_svd}.  It is evident that Med2Meta has been able to learn medically meaningful representations such that they have been separated into distinct clusters compared to the baseline plots. 

\section{Related Works}

\textbf{Meta-Embedding Learning}: Usefulness of meta-embeddings in NLP tasks are realized in some very recent works. \cite{yin2015learning} is one of the first works in meta-embedding learning that proposed a model, 1TON, that learns meta-embeddings by projecting them to source embeddings using separate projection matrices. 1TON is then extended to 1TON+ to account for out-of-vocabulary (OOV) words by first predicting their source embeddings. An unsupervised locally-linear approach is used in \cite{bollegala2017think} to learn meta-embedding of each word based on its local neighborhood of word embeddings. Different variants of autoencoder are used in \cite{bollegala2018learning} to learn meta-embeddings from pre-trained word embeddings.   

Some other works did not learn meta-embeddings, but closely resemble its mechanism by demonstrating effectiveness of integration of different types of embeddings. \cite{ma2018drug} learns similarities between drugs by integrating embeddings learned from a multi-view graph auto-encoder using attention mechanism. A two-sided neural network is used in \cite{luo2014pre} to learn embeddings from multiple data sources.

\section{Conclusion}Leveraging the heterogeneous data types in EHR can be beneficial to learning embeddings that holistically reflect all semantic properties among different medical concepts.  Our proposed approach, Med2Meta, learns feature-specific embeddings using a graph auto-encoder by considering each data type as a separate view. It then models integration of embeddings as a meta-embedding learning problem so that latent similarities and natural clusters between medical concepts are captured in the meta-embedding space through joint reconstruction across all the views. Empirical results on three different tasks and visualization with t-sne plots establish the superior performance and efficacy of Med2Meta over baselines. 

\balance
\bibliographystyle{apalike}
{\small
\bibliography{example}}

\end{document}